\documentclass[sigconf]{acmart}

\usepackage{bbding}
\usepackage{multirow}
\usepackage{subfig}
\usepackage{colortbl}

\AtBeginDocument{%
  }

\copyrightyear{2025}
\acmYear{2025}
\setcopyright{cc}
\setcctype{by}
\acmConference[ICMR '25]{Proceedings of the 2025 International Conference on Multimedia Retrieval}{June 30-July 3, 2025}{Chicago, IL, USA}
\acmBooktitle{Proceedings of the 2025 International Conference on Multimedia Retrieval (ICMR '25), June 30-July 3, 2025, Chicago, IL, USA}
\acmDOI{10.1145/3731715.3733338}
\acmISBN{979-8-4007-1877-9/2025/06}

\begin{document}

\title{Few-Shot Generalized Category Discovery With Retrieval-Guided Decision Boundary Enhancement}

\author{Yunhan Ren}
\email{yhren1218@gmail.com}
\orcid{0009-0008-3356-1978}
\affiliation{
  \institution{Clemson University}
  \city{Clemson}
  \state{South Carolina}
  \country{USA}
}

\author{Feng Luo}
\email{luofeng@clemson.edu}
\orcid{0000-0002-4813-2403}
\affiliation{
  \institution{Clemson University}
  \city{Clemson}
  \state{South Carolina}
  \country{USA}
}

\author{Siyu Huang}
\email{siyuh@clemson.edu}
\orcid{0000-0002-2929-0115}
\affiliation{
  \institution{Clemson University}
  \city{Clemson}
  \state{South Carolina}
  \country{USA}
}

\renewcommand{\shortauthors}{Yunhan Ren, Feng Luo, \& Siyu Huang}

\begin{abstract}
While existing Generalized Category Discovery (GCD) models have achieved significant success, their performance with limited labeled samples and a small number of known categories remains largely unexplored. In this work, we introduce the task of Few-shot Generalized Category Discovery (FSGCD), aiming to achieve competitive performance in GCD tasks under conditions of known information scarcity. To tackle this challenge, we propose a decision boundary enhancement framework with affinity-based retrieval. Our framework is designed to learn the decision boundaries of known categories and transfer these boundaries to unknown categories. First, we use a decision boundary pre-training module to mitigate the overfitting of pre-trained information on known category boundaries and improve the learning of these decision boundaries using labeled samples. Second, we implement a two-stage retrieval-guided decision boundary optimization strategy. Specifically, this strategy further enhances the severely limited known boundaries by using affinity-retrieved pseudo-labeled samples. Then, these refined boundaries are applied to unknown clusters via guidance from affinity-based feature retrieval. Experimental results demonstrate that our proposed method outperforms existing methods on six public GCD benchmarks under the FSGCD setting. The codes are available at: \href{https://github.com/Ryh1218/FSGCD}{https://github.com/Ryh1218/FSGCD}.
\end{abstract}


\begin{teaserfigure}
  \includegraphics[width=\textwidth]{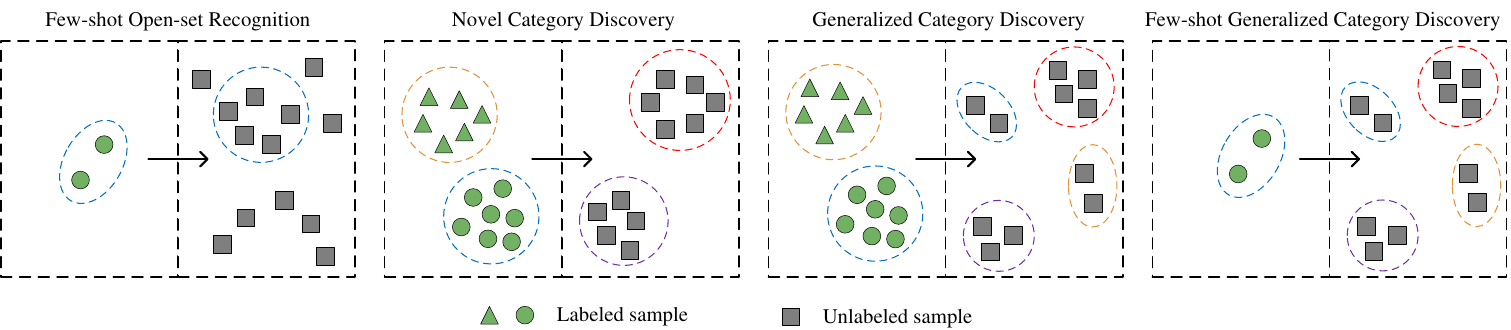}
  \caption{The comparison between the Few-shot Generalized Category Discovery (FSGCD) setting and other representative experimental settings. Unlike Few-shot Open-set Recognition (FSOR), which primarily focuses on distinguishing between known and novel classes, FSGCD is able to categorize unknown samples into specific categories. Compared to Novel Category Discovery (NCD) and Generalized Category Discovery (GCD), FSGCD aims to achieve equal or even superior performance with a limited number of known categories and labeled samples.}
  \Description{}
  \label{fig1}
\end{teaserfigure}


\maketitle

\section{Introduction}
\label{sec:intro}
\begin{figure}[t]
\centering
\includegraphics[width=\linewidth]{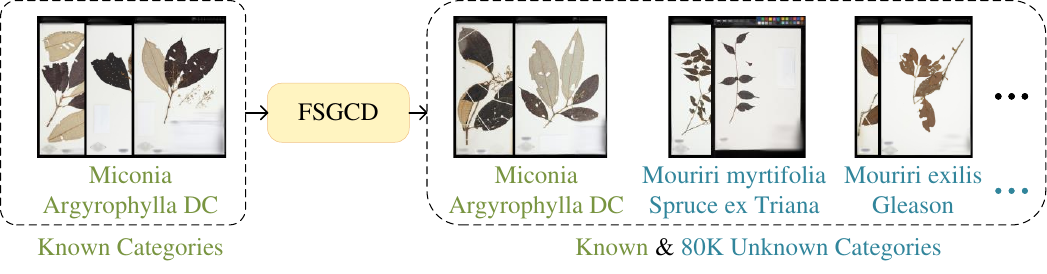}
\caption{An example of Few-shot Generalized Category Discovery (FSGCD) in new plant discovery, which aims at discovering massive (e.g., 80K \cite{tan_herbarium_2019}) different unseen plant species in the real world by using only a few labeled samples.}
\Description{}
\label{Story}
\end{figure}

Generalized Category Discovery (GCD) \citep{vaze_generalized_2022} was recently developed to learn the discriminative representations from labeled samples in known categories and use them to classify unlabeled samples that can belong to either known or novel categories. Existing methods \citep{an_transfer_2023, choi_contrastive_2024, otholt_guided_2024, shi_unified_2024, vaze_generalized_2022, wang_sptnet_2024, wen_parametric_2023, zhao_learning_2023} perform such knowledge transfer under the assumption that there are sufficient labeled samples belonging to ample known categories. However, this assumption is sometimes difficult to satisfy in practice. For instance, in the wildlife monitoring field, sensing devices enable cost-effective data acquisition to generate large amounts of unlabeled data while discovering unknown species is one of the primary goals for zoologists \citep{mou_monitoring_2023}. However, wildlife conservation data are often not publicly available, while obtaining massive labeled data is extremely costly, requiring an effective way to transfer limited known information to classify unknown species \citep{tuia_perspectives_2022}. Similarly, in plant classification, according to \cite{tan_herbarium_2019}, 80,000 plant species are yet to be discovered, and datasets particularly containing these species are difficult to obtain in practice \citep{lu_review_2021}. Figure \ref{Story} illustrates a potential approach to discovering a vast number of new plant species without the need for extensive manual labeling. Furthermore, collecting ample labeled data for certain classes may be expensive due to the stringent requirements for domain expertise or sometimes even infeasible, such as medical instances of rare diseases in dermatology or satellite images of airplane wreckage \citep{adamson_machine_2018, dai_pfemed_2023, feng_learning_2023, guo_broader_2020}. Hence, accessing specific domains or images belonging to sensitive categories may be difficult, limiting the applications of GCD in these scenarios.

This work proposes a new GCD setting: Few-shot Generalized Category Discovery (FSGCD). FSGCD is defined as a GCD task with limited labeled samples and known categories. Typically, the labeled samples and known categories comprise less than 20\% of the total samples and category count. As illustrated in Figure \ref{fig1}, compared to Few-shot Open-set Recognition (FSOR), FSGCD can classify specific novel categories rather than merely rejecting unseen samples as ‘Novel’. Unlike Novel Category Discovery (NCD) and traditional GCD which rely on sufficient known information, FSGCD aims to achieve competitive open-set classification performance in few-shot scenarios. As an image classification problem, GCD seeks to group similar image features together and define cluster boundaries based on these feature groups, using them as references for classification. These cluster boundaries are commonly referred to as 'decision boundaries,' a crucial concept that impacts final classification accuracy. We consider GCD to be a decision boundary transfer problem, where the objective is to train a model that learns the relative relationships of labeled samples and aims to replicate these relationships among unlabeled samples, forming the decision boundaries of category clusters. To make this transfer effective, two prerequisites should be satisfied: 1) the model is able to precisely learn the distinctive decision boundaries of known categories, and 2) the learned decision boundaries carry sufficient information to estimate all the unknown categories’ decision boundaries. However, in the FSGCD setting, the above-mentioned prerequisites are often unable to be satisfied. On the one hand, the limited number of labeled samples makes it challenging for the model to learn accurate decision boundaries for known categories, causing obstacles in classifying unlabeled samples belonging to known categories. On the other hand, the limited proportion of known categories is insufficient to estimate the decision boundaries of all unknown categories.

Consequently, we propose a decision boundary enhancement paradigm incorporating a decision boundary pre-training module and a retrieval-guided decision boundary optimization module to address these two challenges, inspired by \cite{wang_sptnet_2024}. The \textbf{decision boundary pre-training module} aims to fully exploit the limited known information under the FSGCD task in two aspects: preventing the overfitting of pre-trained knowledge and better leveraging the labeled samples. Firstly, for GCD models in original settings such as \cite{choi_contrastive_2024, vaze_generalized_2022, wen_parametric_2023}, they typically fine-tune the last transformer block of the ViT backbone \cite{dosovitskiy_image_2020}. However, under the FSGCD setting, the number of novel categories far exceeds that of known ones, making such schemes prone to overfitting on the decision boundaries of known categories, which hinders generalization to unknown categories. To tackle this issue, we introduce a parametric adapter to reduce the number of learnable parameters while preserving pre-trained decision boundaries. Inspired by \cite{chen_adaptformer_2022}, although adapters for large pre-trained backbones have proven effective in supervised tasks, few studies have demonstrated their effectiveness in open-set unsupervised clustering tasks. We innovatively incorporate it into FSGCD, aiming to learn the decision boundaries of known classes even when labeled samples are limited, and to benefit the estimation of unknown decision boundaries using pre-trained knowledge. Simultaneously, to fully exploit the information contained in labeled samples, our model undergoes exclusive pre-training with labeled samples to capture the data distribution and cluster boundaries of known categories. This phase is referred to as known boundary pre-training since classification boundaries are pre-trained before predicting unlabeled samples.

\begin{figure}[t]
\centering
\includegraphics[width=0.95\linewidth]{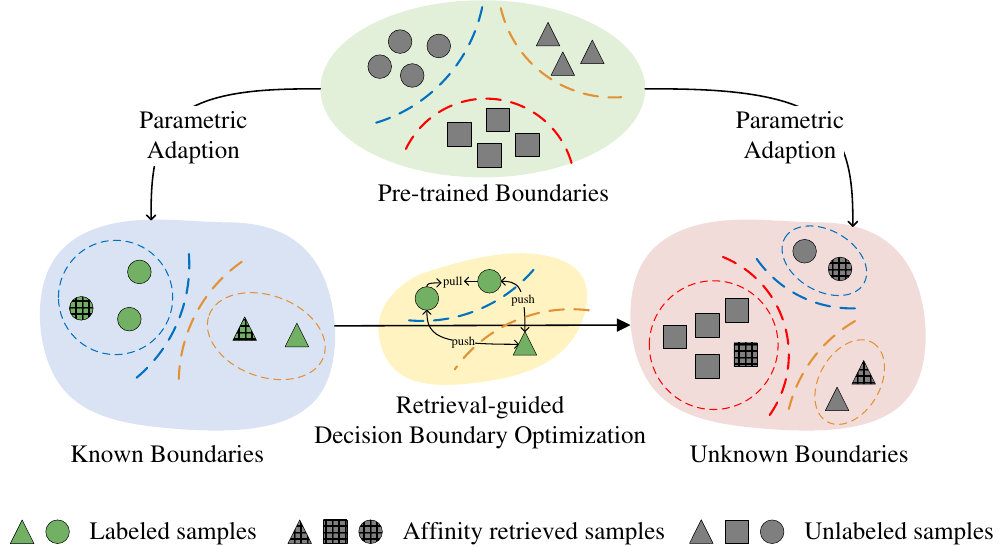}
\caption{The proposed paradigm aims to improve the quality of decision boundaries under the FSGCD setting. By employing the decision boundary pre-training module, pre-trained boundaries are better preserved, thereby compensating for the lack of known knowledge. Subsequently, the retrieval-guided decision boundary optimization module helps the transfer between known and unknown boundaries.}
\label{knowledge}
\Description{}
\end{figure}

On the other hand, traditional GCD \cite{vaze_generalized_2022} relies solely on unsupervised contrastive loss to implicitly form decision boundaries between unlabeled samples, which proves less effective in the FSGCD setting due to the lack of known information and the high proportion of unlabeled samples. In contrast, our goal is to explicitly guide the model to estimate the decision boundaries for unknown categories by referencing the enhanced decision boundaries of known categories through the \textbf{retrieval-guided decision boundary optimization module}. Specifically, we propose a two-stage decision boundary enhancement approach aimed at effectively leveraging improvements in both known and unknown decision boundaries through affinity-based feature retrieval. In the first stage, an affinity-based feature retrieval method is used to identify reliable pseudo-labeled samples, compensating for the scarcity of known information. By employing contrastive learning between labeled samples and their retrieved pseudo-labeled neighbors, our model further learns reliable decision boundaries for known categories. In the second stage, our model transfers the relative relationships among known categories to infer those among unknown categories. This inference is achieved by explicitly simulating the relationships among labeled samples and their affinity-retrieved neighbors that were pre-trained in the first stage, now using unlabeled samples and their affinity-retrieved neighbors. Figure \ref{knowledge} summarizes the explicit decision boundary transfer within the decision boundary pre-training module and the retrieval-guided decision boundary optimization module, illustrating how our model optimizes the quality of both known and unknown decision boundaries. By ensuring that enhanced decision boundaries of known categories can effectively generalize to unknown categories, our model improves overall model performance.

The proposed framework has been evaluated on six public GCD benchmarks \citep{krause_3d_2013, krizhevsky_learning_2009, russakovsky_imagenet_2015, tan_herbarium_2019, wah_caltech-ucsd_2011}. Under the FSGCD setting, our framework achieves significant performance improvements compared to previous state-of-the-art methods. This demonstrates the effectiveness of our proposed decision boundary optimization. Our main contributions can be summarized as follows: 
\begin{itemize} 
\item We introduce the Few-shot Generalized Category Discovery (FSGCD), which addresses the GCD task with a severely limited number of labeled samples and known categories. Along with corresponding baselines and metrics, FSGCD has the potential to advance many real-world category discovery problems. 
\item We present a novel retrieval-guided decision boundary enhancement paradigm for the FSGCD problem, aiming to precisely learn the distinctive decision boundaries of known categories and explicitly improve the quality of novel categories’ decision boundaries through affinity-based feature retrieval.
\item We conduct extensive experiments on multiple GCD benchmarks, demonstrating that our proposed framework significantly outperforms existing state-of-the-art methods in the FSGCD setting. 
\end{itemize}

\section{Related Work}
\label{sec:rw}

\subsection{Few-Shot Learning}
Few-shot learning (FSL) represents a machine learning paradigm defined as learning from limited samples under strong supervision \citep{wang_generalizing_2020}. Existing methods mainly encompass three effective strategies: data augmentation, transfer learning, and meta-learning \citep{song_comprehensive_2023}. \citet{chen_image_2019, li_learning_2021, zhang_adargcn_2021} focus on using data augmentation to enrich known information through pixel transformation, reconstruction, and generation. \citet{alfassy_laso_2019, gao_low-shot_2018, zhang_few-shot_2019} model the known information by mapping pixels into a high-dimensional latent space. Transfer learning seeks to mitigate the severe imbalance between labeled and unlabeled samples by pre-training on extensive datasets and fine-tuning on small-scale target datasets \citep{lu_detect_2024, luo_closer_2023, nakamura_revisiting_2019, song_comprehensive_2023, wang_federated_2022, wang_few-shot_2023}. Meta-learning in FSL learns a meta-tuning model that provides a set of hyperparameters tailored to different tasks upon convergence \citep{song_comprehensive_2023}. \citet{finn_model-agnostic_2017, gupta_look-ahead_2020, jamal_task_2019, rajasegaran_itaml_2020} train hyperparameter generators to address the issue of random initialization.

In this paper, we novelly combine FSL with the practical task of Generalized Category Discover (GCD). We aim to explore the GCD task under conditions of restricted information.

\subsection{Few-Shot Open-Set Recognition}
Few-Shot Open-Set Recognition (FSOR) is proposed to expand the application of FSL in open-set recognition, aiming to recognize positive categories and distinguish them from novel classes with few labeled samples \citep{boudiaf_open-set_2023, deng_learning_2023, song_few-shot_2022, yu_overall_2024, wu_towards_2024, zhang_frequency-aware_2024}. Existing methods can be roughly divided into three categories. The first category uses loss-based methods, where \citet{liu_few-shot_2020} proposes an open-set loss to increase the entropy of classification results for open-set samples, while \citet{wang_glocal_2023} employs a margin-based energy loss to consider both class prototypes and pixel features. The second category utilizes transformation consistency to transform FSOR into a sample-agnostic relative feature transformation problem \citep{jeong_few-shot_2021, zhou_learning_2021}. The third category involves additional open-set information, where \citet{huang_task-adaptive_2022, wang_glocal_2023} employ positive knowledge with the attention mechanism to predict negative prototypes.

Our work extends the FSOR by classifying unlabeled samples into specific categories, which enables the exploration of broader application scenarios.

\subsection{Novel/Generalized Category Discovery}
The task of Novel Category Discovery (NCD) aims to categorize unlabeled samples into novel categories using labeled samples from known classes as guidance \citep{fini_unified_2021, han_automatically_2020, han_learning_2019, zhao_novel_2021}. Building upon NCD, \citet{vaze_generalized_2022} formulates the task of GCD, where some samples in known classes may remain unlabeled, extending the NCD task to more general scenarios. Early works employ both supervised and unsupervised contrastive learning to facilitate knowledge transfer \citep{chen_simple_2020, fei_xcon_2022, khosla_supervised_2020, vaze_generalized_2022}. Many subsequent works \citep{choi_contrastive_2024, wen_parametric_2023} concentrate on improving feature representations. At the same time, \citet{choi_contrastive_2024, zhao_learning_2023} unify contrastive learning and class number estimation within a single framework, further easing the restrictions in the GCD task.

In this work, we introduce the FSGCD task, which aims to explore GCD scenarios where the number of labeled samples and known categories are severely limited.

\section{Method}
\subsection{Few-shot Generalized Category Discovery}
We first formulate the task of GCD. For a partially labeled train set $\mathcal{D}_{t} = \mathcal{D}_{l} \cup \mathcal{D}_{u}$, the labeled subset $\mathcal{D}_{l}$ is defined as $\mathcal{D}_{l} = \left\{\mathbf{x}_{i}, y_{i} \right\}_{i}^{N} \in \mathcal{X}_{l} \times \mathcal{Y}_{l}$, where $\mathcal{X}_{l}$ and $\mathcal{Y}_{l}$ represent the set of all possible labeled images and corresponding known category labels, respectively. Specifically, the known categories set $\mathcal{Y}_{l} = \mathcal{C}_{kwn} \subset \mathcal{C}$, indicating that all labeled samples share known class labels. The unlabeled subset $\mathcal{D}_{u}$ is formulated as $\mathcal{D}_{u} = \left\{\mathbf{x}_{i} \right\}_{i}^M \in \mathcal{X}_{u}$, while its label set $\mathcal{Y}_{u}$ under GCD setting is defined as $\mathcal{Y}_{u} = \mathcal{C} = \mathcal{C}_{kwn} \cup \mathcal{C}_{uwn}$ and cannot be accessed by the model during training, which implies that unlabeled samples can belong to either known or unknown categories. Following \citep{vaze_generalized_2022, wen_parametric_2023}, the task of GCD is to classify the unlabeled samples in $\mathcal{D}_{u}$ into $\vert \mathcal{C} \vert$ clusters using only the information from $\mathcal{D}_{l}$. Following previous works, we assume that the total number of categories $\vert \mathcal{C} \vert$ is known.

\begin{figure*}[t]
\centering
\includegraphics[width=0.95\linewidth]{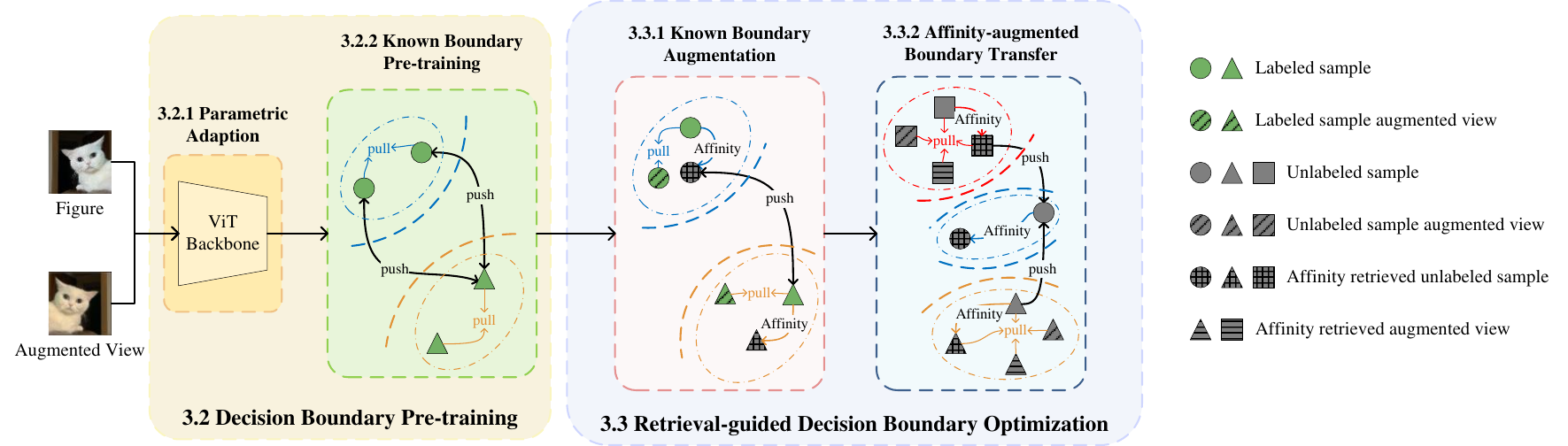}
\caption{An overview of the proposed FSGCD framework. The ViT backbone with parametric adaption is used to extract features from both the original image and its augmented view, as detailed in \ref{pdb}. Then, to fully leverage the known information, decision boundaries of known categories are pre-trained in the Known Boundary Pre-training phase using $\mathcal{D}_{l}$, as described in \ref{kbp}. The two-stage retrieval-guided decision boundary optimization is divided into two steps: refined boundaries are further augmented and then transferred to unknown categories for decision boundary prediction, as described in \ref{ddbo}. Specifically, the Known Boundary Augmentation enhances the model's understanding of known class information through affinity-based data retrieval, as explained in \ref{kba}. Then, the augmented cluster boundaries and known class information are transferred to the novel categories through building similar triplets, as outlined in \ref{abt}.}
\label{Pipeline}
\Description{}
\end{figure*}

For the FSGCD setting, we define the size of the known categories set $\vert \mathcal{C}_{kwn} \vert$ to be much smaller than the total categories set $\vert \mathcal{C}\vert$, formulated as $c_{l} = \vert \mathcal{C}_{kwn} \vert / \vert \mathcal{C} \vert \leq 0.2$. As for the count of labeled samples, we assume that the proportion of labeled samples within the known categories is relatively small, which can be expressed as $p_{l} = \vert \mathcal{X}_{l} \vert / \vert \mathcal{X}_{kwn} \vert \leq 0.2$. Here, $\mathcal{X}_{kwn}$ denotes the set of all samples in known categories $\mathcal{Y}_{l}$, including both labeled and unlabeled ones. The proportion $p_{l}$ represents the ratio of labeled samples $\mathcal{X}_{l}$ to all known category samples $\mathcal{X}_{kwn}$. Figure \ref{Pipeline} summarizes our proposed FSGCD framework, where we enhance the decision boundaries of $\mathcal{C}_{kwn}$ and $\mathcal{C}_{uwn}$ through our proposed decision boundary enhancement paradigm.

\subsection{Decision Boundary Pre-training Module} \label{pdb}
\subsubsection{Parametric Adaption}
We improve the decision boundary representations by preventing the overfitting of pre-trained decision boundaries on downstream known categories. In the FSGCD setting, the limitations of known categories cause a high risk of overfitting and a lack of information to be fine-tuned. To address these issues, we introduce a method that helps retain pre-trained decision boundaries, compensating for known category limitations. Doing so allows us to fine-tune fewer parameters, thus preserving the model’s generalization capability for estimating unknown classes. Specifically, following \citep{chen_adaptformer_2022}, we freeze all the parameters of the ViT backbone and attach a learnable path beside the original MLP. The number of parameters is limited by trainable MLP with low dimensional bottleneck, i.e. adapter, hence preventing overfitting on downstream clustering boundaries. Formally, for an input feature $\mathbf{v}_{i}$, it will pass the learnable path by:
\begin{equation}
\overline{\mathbf{v}}_{i} = \text{ReLU}(\text{LN}(\mathbf{v}_{i})\cdot \mathbf{W}_{down}) \cdot \mathbf{W}_{up}
\end{equation}
Then, with a scale factor $s$, the processed feature will fuse with the output of the original MLP via a residual connection. This process will be formulated as:
\begin{equation}
\mathbf{v}_{i} = \text{MLP}(\text{LN}(\mathbf{v}_{i}')) + \mathbf{v}_{i}' + s \cdot \overline{\mathbf{v}}_{i}
\end{equation}

\paragraph{Insight into decision boundary} To explore the effect of parametric adaption on decision boundaries, we adopt the Calinski-Harabasz index (CH Index) to examine the quality of clustering \cite{calinski_dendrite_1974}. The CH Index indicates the ratio of the between-cluster dispersion to the within-cluster dispersion. A higher CH Index indicates greater similarity between a data point and its centroid and greater distinction between this data point and other clusters. We randomly select 5000 samples from $\mathcal{D}_{u}$ of the CIFAR100 dataset, extract their features using ViT with and without parametric adaption, and calculate the CH index of cluster results, along with their classification accuracy on $\mathcal{C}$, $\mathcal{C}_{kwn}$ and $\mathcal{C}_{uwn}$. As shown in Table \ref{ablation_adapter}, under the FSGCD setting, the parametric adaption leads to more cohesive intra-clusters and more distinctive inter-clusters, resulting in higher classification accuracy on both known and unknown categories.

\begin{table}[!t]
\footnotesize
\centering
    \begin{tabular}{c c c c c c}
    \toprule
        Adapter & Params & CH Index & ALL & OLD & NEW \\
    \midrule
        \XSolidBrush & 7.1M & 24.9 & 58.6 & 71.9 & 58.0 \\
        \Checkmark & \textbf{1.1M} & \textbf{26.2} & \textbf{71.3} & \textbf{73.4} & \textbf{71.2} \\
    \bottomrule
    \end{tabular}
    \caption{Experiment on parametric adaption. The CH Index represents the Calinski-Harabasz index for measuring clustering quality. The ALL, OLD, and NEW represent the classification accuracy on $\mathcal{C}$, $\mathcal{C}_{kwn}$ and $\mathcal{C}_{uwn}$.}
    \label{ablation_adapter}
\end{table}

\subsubsection{Known Boundary Pre-training} \label{kbp}
The model first pre-trains known boundaries to fully exploit the known information before encountering novel categories. This is achieved by constructing triplets of labeled sample features and applying the standard triplet loss. Each triplet is constructed using only labeled features, where the positive is a randomly selected feature from the same category as the anchor, and the negative is randomly selected from a different category. By optimizing this triplet loss, the model will learn to distinguish between known categories, which serve as a reference for predicting novel decision boundaries.

\paragraph{Known Triplet Loss} Standard triplet loss \citep{schroff_facenet_2015} can be formulated as: 
\begin{equation}
    \mathcal{L}_{tl}^{i}(\mathbf{a}_{i}, \mathbf{p}_{i}, \mathbf{n}_{i}, \alpha) = \text{max}(\vert\vert \mathbf{a}_{i} - \mathbf{p}_{i}\vert\vert_{2}^{2} - \vert\vert \mathbf{a}_{i} - \mathbf{n}_{i}\vert\vert_{2}^{2} + \alpha, 0) \label{triplet:single}
\end{equation}
where $\mathbf{a}_{i}, \mathbf{p}_{i}, \mathbf{n}_{i}$ represent the anchor, positive and negative embeddings, respectively; $\alpha$ denotes the margin between positive and negative pairs. For optimization, the triplet losses of minibatch $b$ can be formulated as:
\begin{equation}
    \mathcal{L}_{tl} = \frac{1}{\vert b \vert} \sum_{i=1}^{\vert b \vert} \mathcal{L}_{tl}^{i}(\mathbf{a}_{i}, \mathbf{p}_{i}, \mathbf{n}_{i}, \alpha) \label{triplet:batch}
\end{equation}

In the FSGCD framework, the known boundary is pre-trained using standard triplet loss on labeled sample features as shown in Equations \ref{triplet:single} and \ref{triplet:batch}. Specifically, for a minibatch $b$ consisting of only the labeled features $\left\{\mathbf{v}_{i}^{l} \right\}_{i=1}^{\vert b \vert }$, its known triplet loss is expressed as:
\begin{equation}
    \mathcal{L}_{k} = \frac{1}{\vert b \vert} \sum_{i=1}^{\vert b \vert} \mathcal{L}_{tl}(\mathbf{v}_{i}^{l}, \tilde{\mathbf{v}}_{i} \in \left\{ \mathcal{P}(\mathbf{v}_{i}^{l}) \right\} , \tilde{\mathbf{v}}_{i} \notin\left\{ \mathcal{P}(\mathbf{v}_{i}^{l}) \right\}) \label{Lk}
\end{equation}
where $\tilde{\mathbf{v}_{i}}$ represents a randomly selected sample inside $b$ and $\left\{ \mathcal{P}(\mathbf{v}_{i}^{l}) \right\}$ denotes the set of all samples belonging to the same category as $\mathbf{v}_{i}^{l}$ within the minibatch $b$. By optimizing this triplet loss, the model learns the decision boundaries of known categories, which are then used as a reference for later transferring.

\subsection{Retrieval-guided Decision Boundary Optimization} \label{ddbo}
The known boundary pre-training has utilized the known information to enhance the learning of known categories' decision boundaries. After this, we propose the Retrieval-guided Decision Boundary Optimization to further retrieve the unlabeled samples in known categories and use such information to infer the decision boundaries of novel categories.

\subsubsection{Known Boundary Augmentation} \label{kba}
After the pre-training of known categories relationships, we facilitate further learning of known decision boundaries by enriching the limited known knowledge. Under the FSGCD setting, the scarcity of labeled samples hinders learning precise feature representation and discriminative cluster boundaries. Therefore, we incorporate a simple affinity-based retrieval approach to introduce pseudo-labeled features from unlabeled data, thereby providing extra samples for representation learning. We then utilize affinity supervised loss on both labeled and pseudo-labeled samples to learn robust feature representations and precise decision boundaries.

\paragraph{Affinity Supervised Loss} For each labeled feature $\mathbf{v}_{i}^{l} \in \left\{ \mathbf{v}^{l} \right\}$, we compute the nearest neighbor based on feature affinity, denoted as $\mathbf{v}_{i}^{a} = f_{a}(\mathbf{v}_{i}^{l})$, where $f_{a}(\cdot)$ indicates the affinity function. Then, $\mathbf{v}_{i}^{a}$ will be labeled with $y_{i}$, indicating the incorporation of additional knowledge. Consequently, the new set of labeled features can be represented as $\left\{ \mathbf{v}^{al} \right\} = \left\{ \mathbf{v}^{l} \right\} \cup \left\{ \mathbf{v}^{a} \right\}$. We calculate supervised contrastive loss on both labeled and affinity-augmented features \citep{khosla_supervised_2020, vaze_generalized_2022}. Specifically, this supervised loss guides the model in learning accurate and clear decision boundaries by drawing intra-class features closer together. Formally, for each labeled or affinity-augmented feature $\mathbf{v}_{i}^{al} \in \left\{ \mathbf{v}^{al}_{b} \right\}$ in minibatch $b$, its supervised loss is defined as:
\begin{equation}
    \mathcal{L}_{s}^{(i)} = -\frac{1}{\vert \left\{ \mathcal{P} \right\} \vert}\sum_{\mathbf{v}_{p} \in \left\{ \mathcal{P} \right\}} \log \frac{\exp (\mathbf{v}_{i}^{al} \cdot \mathbf{v}_{p} / \tau_{s})} {\sum_{\mathbf{v}_{n} \notin \left\{ \mathcal{P} \right\}} \exp (\mathbf{v}_{i}^{al} \cdot \mathbf{v}_{n} / \tau_{s})}
\end{equation}
where $\left\{ \mathcal{P} \right\} = \left\{ \mathcal{P}(\mathbf{v}_{i}^{al}) \right\}$ indicates the set of features that share the same class label $y_{i}$ with $\mathbf{v}_{i}^{al}$ within the minibatch $b$. $\tau_{s}$ indicates temperature hyperparameter. Then, for minibatch $b$, the supervised loss will be executed for each labeled or affinity-augmented feature, which indicates the minibatch’s \textbf{A}ffinity \textbf{S}upervised \textbf{L}oss can be expressed as:
\begin{equation}
    \mathcal{L}_{asl} = \frac{1}{\vert \left\{ \mathbf{v}^{al}_{b} \right\}\vert} \sum_{i=1}^{\vert\left\{ \mathbf{v}^{al}_{b} \right\}\vert} \mathcal{L}_{s}^{(i)}
\end{equation}

\subsubsection{Affinity-augmented Boundary Transfer} \label{abt}
\paragraph{Knowledge Transfer Loss} The enhanced decision boundaries are acknowledged as one of the optimization objectives, specifically, the knowledge transfer loss. As detailed in Section \ref{kbp}, our model learns the accurate decision boundaries for known categories through triplets of labeled samples, where two intra-class samples form positive pairs and two inter-class samples constitute negative pairs. By creating similar triplets from unlabeled samples and applying triplet loss, the model can leverage the learned relationships among labeled samples to infer relationships between unlabeled samples, essentially transferring the relationships of known category clusters to analogous unlabeled clusters. Specifically, for each unlabeled anchor feature, we identify its affinity-retrieved feature as positive and randomly select another distinct feature from the same minibatch to serve as the negative. Formally, for $b^u$ unlabeled features $\left\{ \mathbf{v}_{i}^{u} \right\}_{i=1}^{b^u}$ in minibatch $b$, the \textbf{K}nowledge \textbf{T}ransfer \textbf{L}oss can be formulated as:
\begin{equation}
    \begin{aligned}
    \mathcal{L}_{ktl} = \frac{1}{b^{u}} \sum_{i=1}^{b^{u}} \mathcal{L}_{tl}(\mathbf{v}_{i}^{u}, \mathbf{v}_{i}^{a}, \tilde{\mathbf{v}}_{i}^{u}), && \text{where} \ \tilde{\mathbf{v}}_{i}^{u} \neq \mathbf{v}_{i}^u
    \end{aligned}
\end{equation}
where $\mathbf{v}_{i}^{a} = f_{a}(\mathbf{v}_{i}^{u})$ denotes the feature most similar to $\mathbf{v}_{i}^{u}$ based on affinity retrieval. $\tilde{\mathbf{v}}_{i}^{u}$ represents a randomly picked feature in minibatch $b$ that is distinct from $\mathbf{v}_{i}^{u}$.

\paragraph{Affinity Loss} We also propose an affinity loss that aims to optimize the affinity-based retrieval process in the above losses to precisely identify the closest affinity feature. We obtain an augmented view of each image by applying cropping, flipping, and color jittering \cite{vaze_generalized_2022}. Consequently, each extracted feature is paired with a unique augmented feature that shares the same semantic meaning, albeit with slightly different deformable information. Then, for each unlabeled feature $\mathbf{v}_{i}^{u} \in \left\{ \mathbf{v}^u \right\}$ and its affinity-retrieved feature $\mathbf{v}_{i}^{a} = f_{a}(\mathbf{v}_{i}^{u})$, we use cosine similarity to pull them closer to each others' augmented features $\mathbf{v}_{i}^{u'}$ and $\mathbf{v}_{i}^{a'}$. Formally, the \textbf{A}ffinity \textbf{L}oss for minibatch $b$, which consists of $b^{u}$ unlabeled features, can be expressed as:
\begin{equation}
\mathcal{L}_{al} = \frac{1}{2 b^{u} } \sum_{i=1}^{ b^{u} } [(1 - \text{sim}(\mathbf{v}_{i}^{u}, \mathbf{v}_{i}^{a'}))
+ (1 - \text{sim}(\mathbf{v}_{i}^{u'}, \mathbf{v}_{i}^{a}))]
\end{equation}
where $\text{sim}(\cdot)$ indicates standard cosine similarity.

\paragraph{Unsupervised Contrastive Loss} Following \citep{vaze_generalized_2022}, we implement an unsupervised contrastive loss to emphasize the distinctiveness of each feature. Formally, for minibatch $b$, the \textbf{U}nsupervised \textbf{C}ontrastive \textbf{L}oss can be expressed as the following formula:
\begin{equation}
    \mathcal{L}_{ucl} = - \frac{1}{\vert b \vert} \sum_{i=1}^{\vert b \vert} \log \frac{\exp(\mathbf{v}_{i} \cdot \mathbf{v}_{i}' / \tau_{u})}{\sum_{j \neq i} \exp(\mathbf{v}_{i} \cdot \mathbf{v}_{j} / \tau_{u})}
\end{equation}
where $\tau_{u}$ denotes the temperature hyperparameter. $\mathbf{v}_{i}'$ represents the augmented view of feature $\mathbf{v}_{i}$ while $\mathbf{v}_{j}$ (where $j \neq i$) denotes a distinct feature within the minibatch $b$, different from $\mathbf{v}_{i}$.

\subsection{Overall Learning Objectives} For the known boundary pre-training stage, Equation \ref{Lk} is used as supervision loss. During the Retrieval-guided Decision Boundary Optimization, the overall learning objective is defined as:
\begin{equation}
    \mathcal{L} = \lambda \cdot \mathcal{L}_{asl} + (1 - \lambda) \cdot \mathcal{L}_{ucl} + \mathcal{L}_{ktl} + \mathcal{L}_{al}
\end{equation}
where $\lambda$ denotes a balancing hyperparameter.

\section{Experiments}

\subsection{Datasets}
We evaluate our proposed model under the FSGCD setting using six image classification datasets: CIFAR10, CIFAR100 \citep{krizhevsky_learning_2009}, ImageNet100 \citep{russakovsky_imagenet_2015}, CUB-200-2011 \citep{wah_caltech-ucsd_2011}, Stanford Cars \citep{krause_3d_2013}, and Herbarium19 \citep{tan_herbarium_2019}. Since the number of known categories is limited under the FSGCD setting, we choose the first $\vert \mathcal{C}_{kwn} \vert$ classes as known categories for each dataset instead of using the splits defined in the Semantic Shift Benchmark (SSB) \citep{vaze_generalized_2022, vaze_open-set_2021}.

In the FSGCD setting, the data splits are carefully considered. For the proportion of known categories ($c_{l}$), 0.05 is set for CIFAR100, while 0.1 is defined for ImageNet100, CUB-200-2011, Stanford Cars, and Herbarium19. For CIFAR10, the proportion of known categories is set to 0.2 to ensure diversity in the known information. The labeled sample proportion ($p_{l}$) is set to 0.1 for CIFAR10, CIFAR100, ImageNet100, Herbarium19 and to 0.2 for CUB-200-2011 and Stanford Cars. Detailed data splits are provided in Table \ref{datasize}.

\begin{table}[t]
\centering
    \resizebox{\columnwidth}{!}{%
    \begin{tabular}[t]{c c c c c c c}
    \toprule
        & CIFAR10 & CIFAR100 & ImageNet100 & CUB & SCars & Herb 19 \\
    \midrule
        $|\mathcal{Y_{L}}|$ & 2 & 5 & 10 & 10 & 10 & 33 \\
        $|\mathcal{Y_{U}}|$ & 10 & 100 & 100 & 200 & 196 & 683 \\
    \midrule
        $|\mathcal{D_{L}}|$ & 1,000 & 250 & 1,297 & 60 & 81 & 207 \\
        $|\mathcal{D_{U}}|$ & 49,000 & 49,750 & 125,818 & 5,934 & 8,603 & 34,108 \\
    \bottomrule
    \end{tabular}%
    }
\caption{
    Experimental settings of FSGCD on six datasets.
}
\label{datasize}
\end{table}%

\definecolor{mygray}{gray}{.9}
\begin{table*}[!t]
    \resizebox{2\columnwidth}{!}{
    \begin{tabular}[b]{p{0.1\textwidth} c ccc ccc ccc ccc ccc ccc}
        \toprule
            \multirow{2.5}{*}{Method} &
            \multicolumn{3}{c}{CIFAR10} &
            \multicolumn{3}{c}{CIFAR100} &
            \multicolumn{3}{c}{ImageNet100} &
            \multicolumn{3}{c}{CUB} &
            \multicolumn{3}{c}{Stanford Cars} &
            \multicolumn{3}{c}{Herbarium 19} \\
            \cmidrule(lr){2-4} \cmidrule(lr){5-7} \cmidrule(lr){8-10}
            \cmidrule(lr){11-13} \cmidrule(lr){14-16} \cmidrule(lr){17-19}
            & ALL & OLD & NEW & ALL & OLD & NEW & ALL & OLD & NEW
            & ALL & OLD & NEW & ALL & OLD & NEW & ALL & OLD & NEW\\
        \midrule
            GCD* & 84.9 & 92.0 & 83.3 & 54.6 & 50.7 & 54.8 & 75.3 & \textcolor{blue}{86.2} & 74.2 & \textcolor{blue}{29.7} & \textcolor{blue}{34.1} & \textcolor{blue}{29.5} & \textcolor{blue}{11.1} & \textcolor{blue}{19.0} & \textcolor{blue}{10.7} & \textcolor{blue}{15.0} & \textcolor{blue}{16.8} & \textcolor{blue}{14.9} \\
            SimGCD & 68.9 & 68.3 & 69.0 & 18.0 & \textcolor{blue}{64.8} & 15.8 & 62.1 & 74.8 & 60.8 & 22.9 & 20.0 & 23.0 & 8.7 & 7.0 & 8.8 & 13.0 & 11.7 & 13.1 \\
            CMS* & \textcolor{blue}{86.0} & \textcolor{blue}{94.6} & \textcolor{blue}{84.0} & \textcolor{blue}{59.7} & 43.1 & \textcolor{blue}{60.5} & \textcolor{blue}{79.2} & 75.0 & \textcolor{blue}{79.7} & 26.6 & 24.5 & 26.7 & 10.8 & \textcolor{red}{\textbf{19.6}} & 10.5 & 11.4 & \textcolor{red}{\textbf{19.4}} & 10.9 \\
            \rowcolor{mygray}
            Our & \textcolor{red}{\textbf{95.0}} & \textcolor{red}{\textbf{97.4}} & \textcolor{red}{\textbf{94.5}} & \textcolor{red}{\textbf{71.3}} & \textcolor{red}{\textbf{73.4}} & \textcolor{red}{\textbf{71.2}} & \textcolor{red}{\textbf{81.1}} & \textcolor{red}{\textbf{92.5}} & \textcolor{red}{\textbf{79.9}} & \textcolor{red}{\textbf{44.7}} & \textcolor{red}{\textbf{45.4}} & \textcolor{red}{\textbf{44.6}} & \textcolor{red}{\textbf{12.3}} & 15.0 & \textcolor{red}{\textbf{12.2}} & \textcolor{red}{\textbf{15.5}} & 14.7 & \textcolor{red}{\textbf{15.6}} \\
        \bottomrule
    \end{tabular}
    }
    \caption{Performances of GCD models under FSGCD setting. *: Evaluation metrics are adapted with other models.}
    \label{FSResults}
\end{table*}%

\subsection{Implementation Details}
We utilize the ViT-B/16 model pre-trained with DINO \citep{caron_emerging_2021, dosovitskiy_image_2020} on ImageNet-1K \citep{russakovsky_imagenet_2015} as the backbone and projection head of our proposed model.
To conserve computational resources, the output dimension of the projection head is manually set to 256. All experiments are conducted on a single RTX-3090 GPU with a batch size of 32. To update the affinity-based nearest neighbor retrieval, the features are pre-extracted before each epoch using only the ViT adapter. The nearest neighbors and pseudo-labels are recorded using an affinity matrix. We set $\tau_{s} = 0.07$, $\tau_{u} = 1.0$ and $\lambda = 0.35$, following \cite{vaze_generalized_2022}. An SGD optimizer is employed with a weight decay of 0.00005, momentum of 0.9, and a learning rate of 0.1. For simplicity and due to the lack of known categories as initial clusters, we use plain K-means clustering instead of the semi-supervised K-means typically employed in traditional GCD settings.

\subsection{Evaluation Protocol}
Following previous works, we utilize the Hungarian algorithm \cite{wright_speeding_1990} to map the clustering results to the ground truth labels through an optimal assignment. We calculate the classification accuracy of samples in $\mathcal{C}, \mathcal{C}_{kwn}$ and $\mathcal{C}_{uwn}$, referred to as ALL, OLD, and NEW, respectively. Following the experiment setting of \cite{wen_parametric_2023}, we report the models' best results on $D_{u}$. Given that the known information is severely limited, and most samples belong to $\mathcal{C}_{uwn}$, for each FSGCD experiment, we report the highest NEW result. Following \cite{vaze_generalized_2022}, we assume the overall number of categories $\vert \mathcal{C} \vert$ is known.

\subsection{Main Results}
Table \ref{FSResults} summarizes the performance of different models under FSGCD setting, including the task baseline \cite{vaze_generalized_2022} (marked as GCD), previous state-of-the-art models \cite{choi_contrastive_2024, wen_parametric_2023} (marked as SimGCD and CMS), and our proposed model. We use the color red to mark the optimal results and blue to indicate the sub-optimal ones. In general, our proposed FSGCD framework significantly improves NEW performance on all the datasets by 11.5\%, 11.2\%, 0.2\%, 15.1\%, 1.5\%, and 0.7\% compared to the second-best model. Regarding the ALL performance, our model also shows significant improvements, achieving gains of 9\%, 11.6\%, 1.9\%, 15\%, 1.2\%, and 0.5\% on CIFAR10, CIFAR100, ImageNet100, CUB, Stanford Cars, and Herbarium 19, respectively. As for the OLD results, our model outperforms the second-best models by 2.8\%, 8.6\%, 6.3\%, and 11.3\% on CIFAR10, CIFAR100, ImageNet100, and CUB, respectively. These results indicate the merit of our proposed FSGCD setting since accurately classifying unlabeled samples with limited known information is challenging, and our proposed decision boundary enhancement paradigm is able to handle this issue effectively. However, our model is less competitive on the OLD performances of Stanford Cars and Herbarium 19. We attribute the decline to the domain gap between pre-trained decision boundaries and downstream category relationships. This gap significantly influences the pre-training and augmentation of known boundaries, leading to lower OLD performance. Despite this, our proposed model effectively generalizes learned information to novel categories, resulting in superior NEW results.

\subsection{Ablation Study}
\paragraph{Contributions of Components} As shown in Table \ref{ablation}, the first row presents the performance of our baseline model on the CIFAR100 dataset. The second row highlights the performance improvements brought by the parametric adaption, i.e., the ViT adapter, demonstrating that the pre-trained decision boundaries are well preserved, thereby preventing overfitting on known categories. The third row shows that the additional enhancement from known boundary pre-training effectively refines the classification boundaries of known categories and provides reliable references for estimating unknown category boundaries, resulting in improved performance on both old and novel categories. The fourth row indicates that the retrieval-guided decision boundary optimization module, which includes known boundary augmentation and affinity-augmented boundary transfer, not only captures relationships between known boundaries but also applies these relationships to unknown classes effectively, demonstrating its superior ability to generalize limited information about unknown decision boundaries. Finally, the fifth row suggests that affinity loss aids in retrieving features' affinity information, ensuring the model learns more robust and precise feature representations. The best results and sub-optimal results are marked in red and blue, respectively.

\begin{table}[t]
\footnotesize
\centering
    \begin{tabular}{c c c c c c c}
    \toprule
        \multirow{2.5}{*}{PA} & \multirow{2.5}{*}{KDP} & \multirow{2.5}{*}{RBD} & \multirow{2.5}{*}{AL} & \multicolumn{3}{c}{CIFAR100}\\ 
        \cmidrule(lr){5-7}
        & & & & ALL & OLD & NEW\\ 
    \midrule
        \XSolidBrush & \XSolidBrush & \XSolidBrush & \XSolidBrush & 54.6 & 50.7 & 54.8 \\ 
        \Checkmark & \XSolidBrush & \XSolidBrush & \XSolidBrush & 59.5 & 55.6 & 59.7\\ 
        \Checkmark & \Checkmark & \XSolidBrush & \XSolidBrush & 65.4 & 63.3 & 65.5 \\ 
        \Checkmark & \Checkmark & \Checkmark & \XSolidBrush & \textcolor{blue}{70.9} & \textcolor{blue}{69.6} & \textcolor{blue}{70.9} \\ 
        \Checkmark & \Checkmark & \Checkmark & \Checkmark & \textcolor{red}{\textbf{71.3}} & \textcolor{red}{\textbf{73.4}} & \textcolor{red}{\textbf{71.2}} \\ 
    \bottomrule
    \end{tabular}
    \caption{Ablation study on investigating various components of our proposed approach. \textbf{PA}: Parametric adaption; \textbf{KDP}: Known boundary pre-training; \textbf{RBD}: Retrieval-guided decision boundary optimization; \textbf{AL}: Affinity loss.}
    \label{ablation}
\end{table}

\begin{figure*}[t]
    \centering
    \subfloat[$c_{l} = 0.1$]{
        \includegraphics[width=0.35\textwidth]{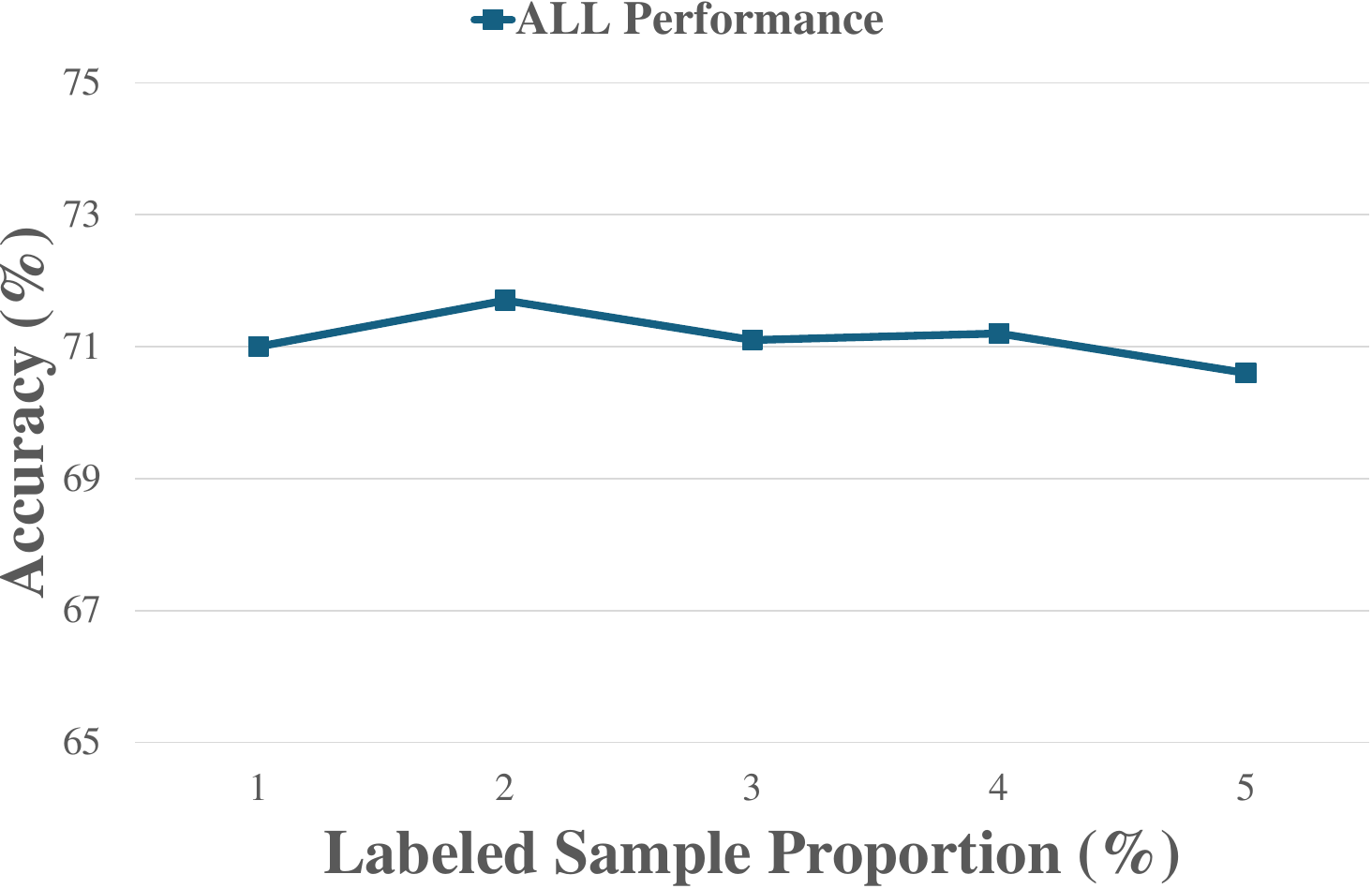}
        \label{ifsa}
    }
    \hspace{0.02\textwidth}
    \subfloat[$p_{l} = 0.1$]{
        \includegraphics[width=0.35\textwidth]{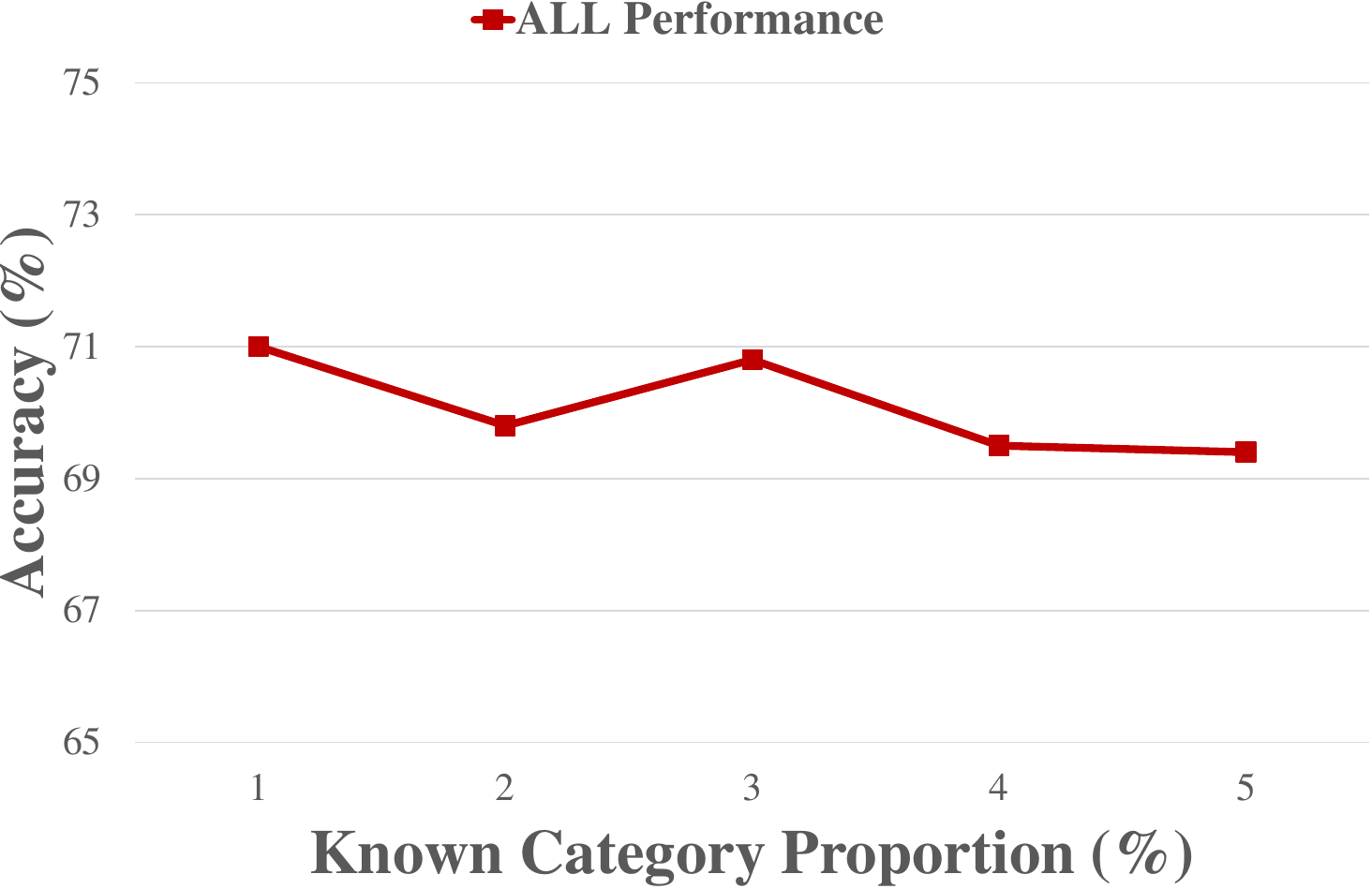}
        \label{ifsb}
    }
    \caption{The ablation study of known categories number and labeled samples proportion under FSGCD setting.}
    \label{ifs}
    \Description{}
\end{figure*}

\paragraph{Ablation of Few-shot Setting} As shown in Figure \ref{ifs}, we explore the influencing factors in the FSGCD setting. In Figure \ref{ifsa}, we fix the known category count $\vert \mathcal{C}_{kwn} \vert$ to 5 and examine the performance of our proposed model under different proportions of labeled samples $p_{l}$. It can be observed that the overall performance of the FSGCD model remains stable, with a slight decline as $p_{l}$ increases. This stability indicates that the FSGCD model is suitable for various extreme scenarios, while the decline highlights that continually enriching known information may lead to overfitting. In Figure \ref{ifsb}, we test whether an increasing number of $\vert \mathcal{C}_{kwn} \vert$ affects our proposed model when $p_{l}$ is fixed at 0.1. The results share similar trends with the previous experiment, further emphasizing the robustness of our FSGCD model. On the other hand, performance declines when there are more known categories while maintaining the labeled sample count, indicating the sensitivity of affinity-based boundary augmentation due to the relatively simple affinity schema used in known boundary augmentation. These experiments demonstrate that our proposed retrieval-guided decision boundary enhancement framework excels in the FSGCD experimental setting and exhibits robustness across various scenarios.

\begin{figure*}[t]
    \centering
    \subfloat[GCD]{
        \includegraphics[width=0.25\textwidth]{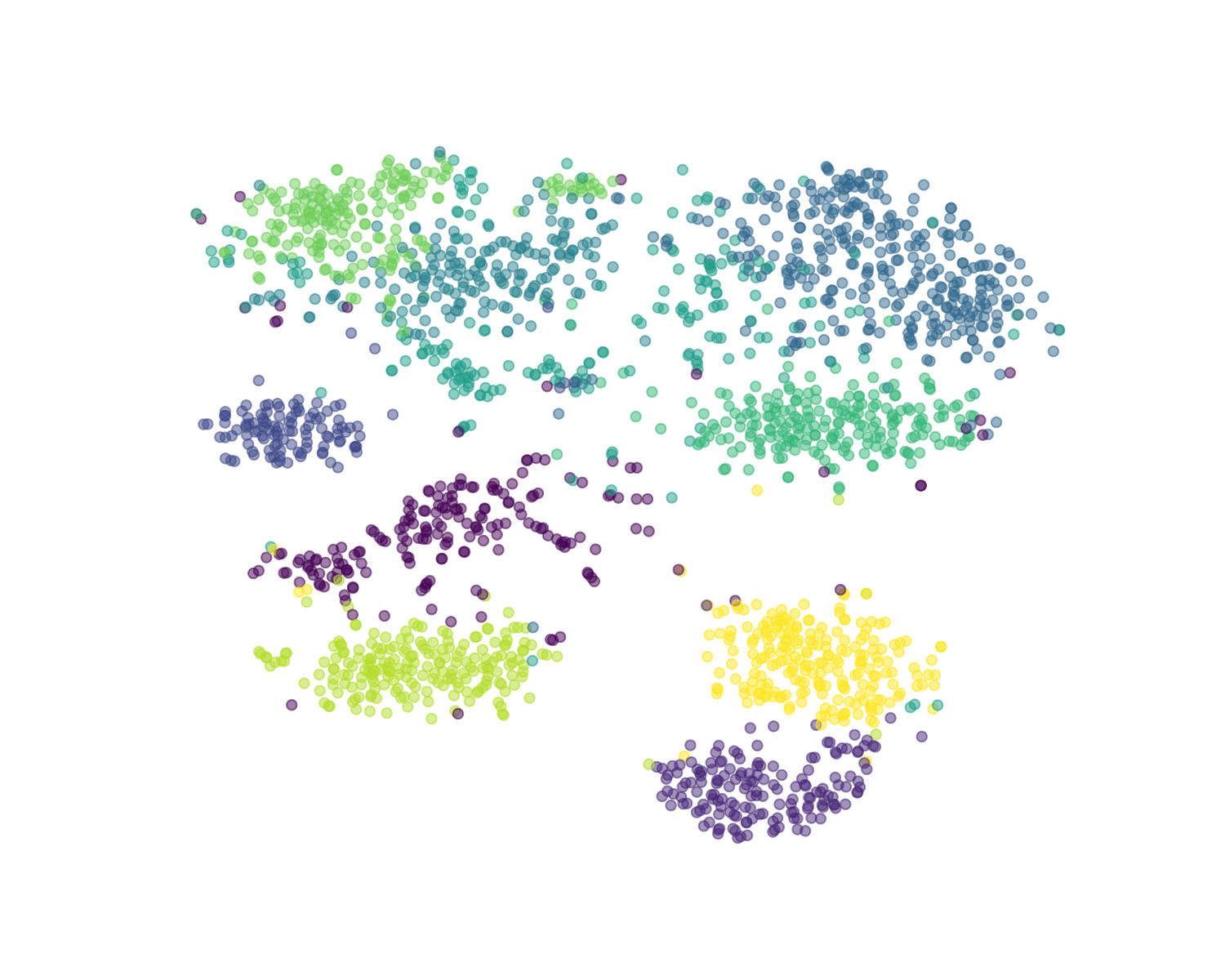} 
        \label{tsne_1}
    }
    \subfloat[CMS]{
        \includegraphics[width=0.25\textwidth]{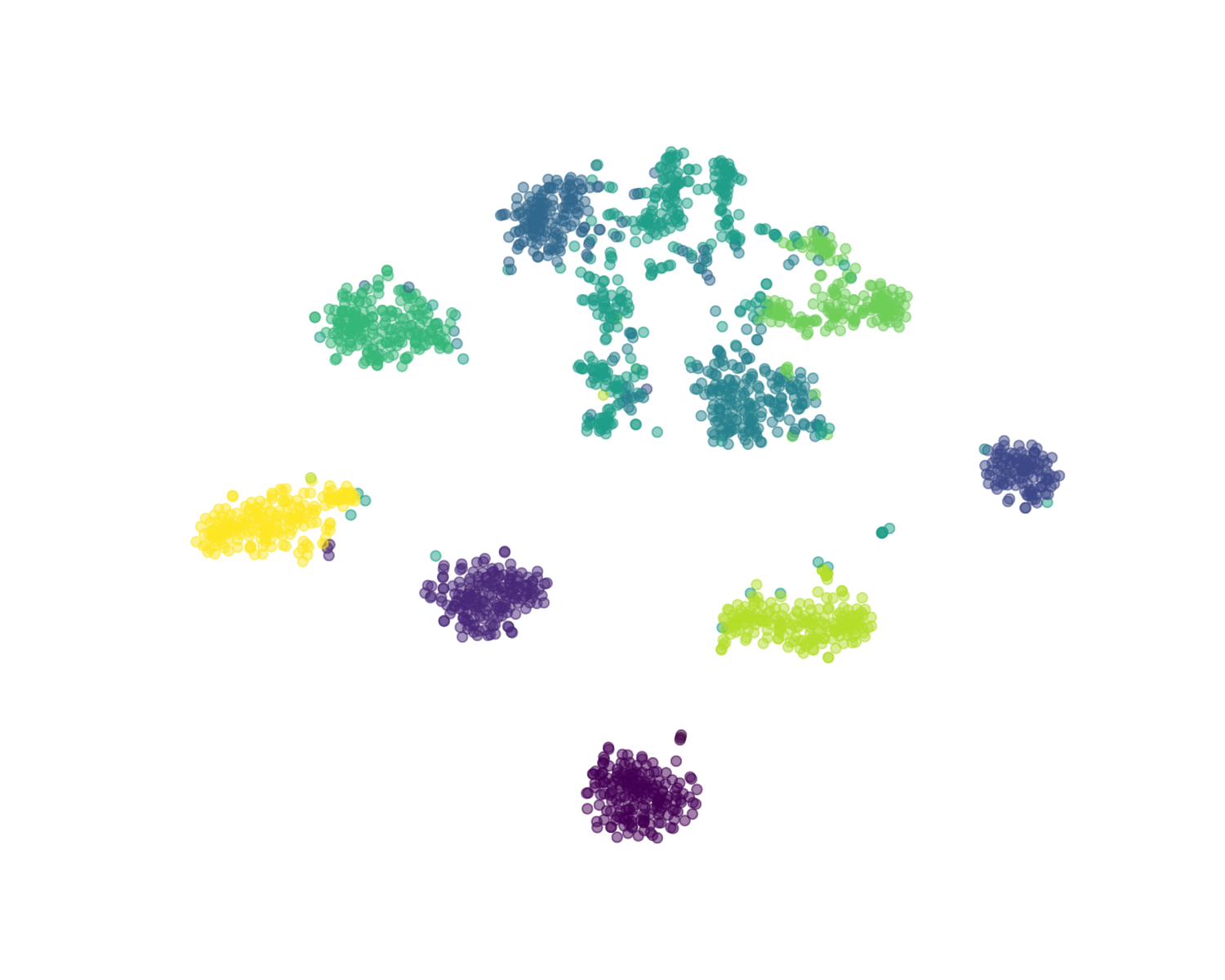} 
        \label{tsne_2}
    }
    \\
    \subfloat[FSGCD without parametric adaption]{
        \includegraphics[width=0.25\textwidth]{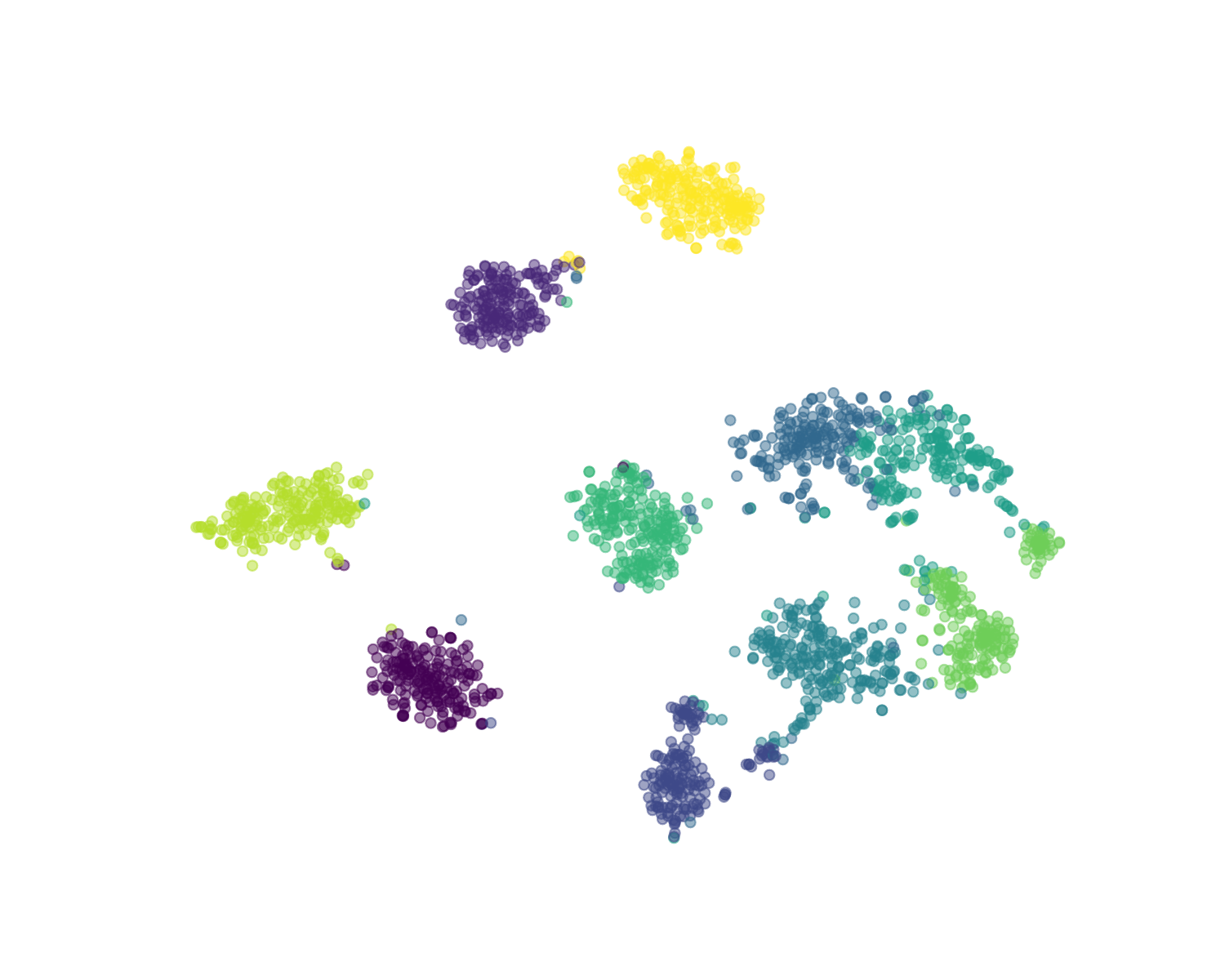} 
        \label{tsne_3}
    }
    \subfloat[FSGCD]{
        \includegraphics[width=0.25\textwidth]{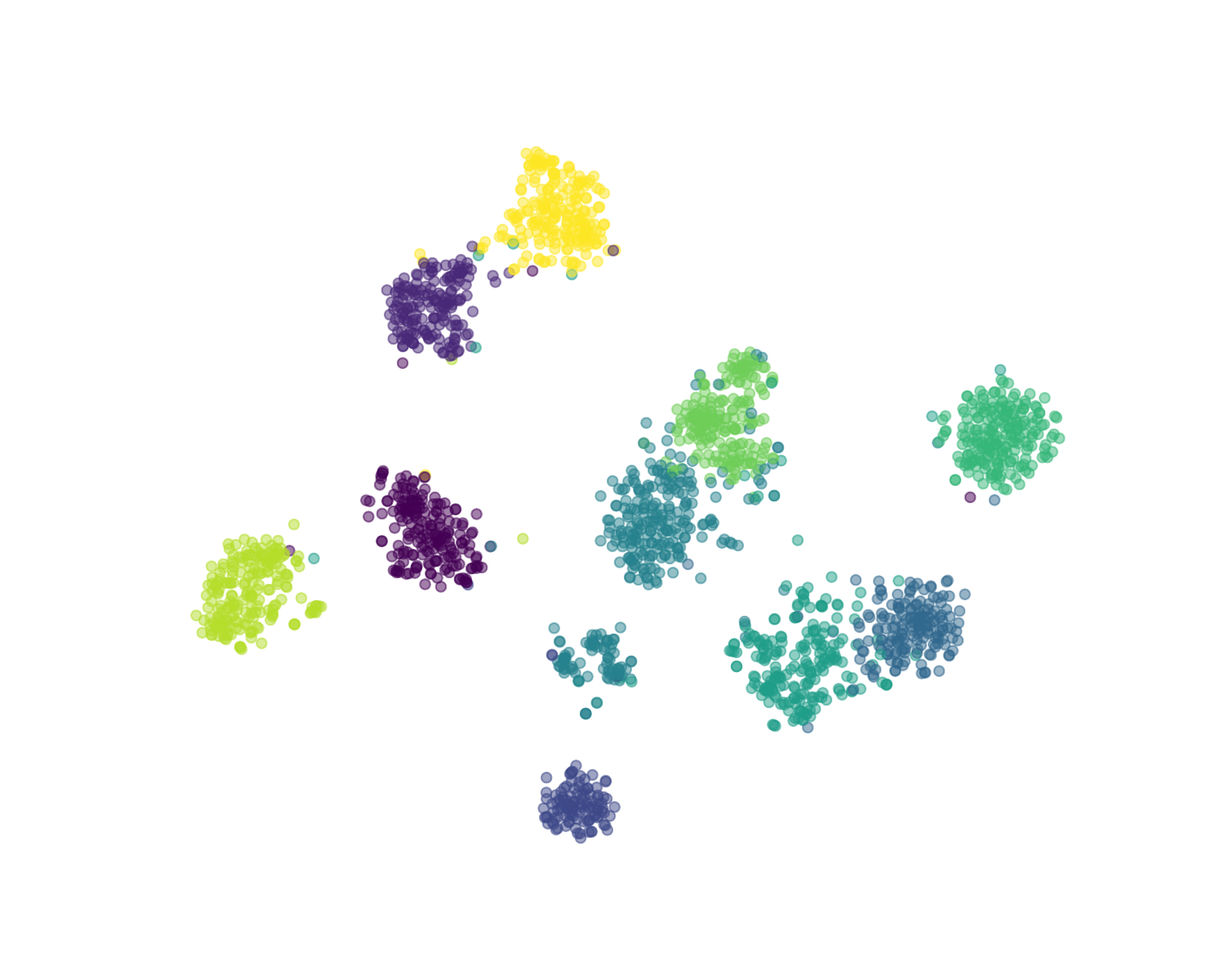} 
        \label{tsne_4}
    }
    \\
    \subfloat{
        \includegraphics[width=0.5\textwidth]{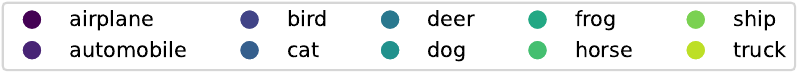} 
        \label{tsne_5}
    }
    \caption{t-SNE visualization of the clustering results for 4 selected models on CIFAR10. Our proposed FSGCD framework results in more distinctive and clear decision boundaries of clustering results.}
    \Description{}
    \label{tsne}
\end{figure*}

\paragraph{Visualization Results} We visualize the clustering results of the original GCD, CMS, FSGCD without parametric adaption and with the known boundary pre-training and the retrieval-guided decision boundary optimization and FSGCD with all the components. Specifically, we conduct experiments on the CIFAR10 dataset and randomly select 2,000 extracted features for visualization using t-SNE \citep{maaten_visualizing_2008}. Figure \ref{tsne} demonstrates the above results. The visualized results show that our framework is able to make the clusters tighter and the category boundaries clearer. This signifies that the proposed methods are capable of grouping similar samples into discriminative clusters, regardless they are labeled or not.

\section{Conclusion}
We have proposed the Few-shot Generalized Category Discovery (FSGCD) task as a challenging extension to GCD, with stricter limitations on both the number of known categories and the availability of labeled samples. To address the information scarcity in FSGCD, we propose a novel decision boundary enhancement paradigm specifically designed to enhance the decision boundaries for both known and unknown categories. Our framework prevents the overfitting of pre-trained decision boundaries on known categories and better exploits the relationships between labeled samples through the decision boundary pre-training module. We also introduce a retrieval-guided decision boundary optimization module, which improves the decision boundaries of known categories by generating affinity-retrieved pseudo-samples and enhances those of unknown categories by leveraging information from learned known cluster boundaries. These approaches collectively tackle the challenges posed by the large proportion of unlabeled samples belonging to unknown categories, enabling more accurate clustering and classification. Experimental evaluations across multiple FSGCD benchmarks demonstrate that our framework significantly outperforms existing state-of-the-art models. The results suggest that our framework not only provides a robust baseline for FSGCD but also offers valuable insights for future research in few-shot open-set recognition tasks with limited known information.

\begin{acks}
This work is partially supported by the AIM for Composites, an Energy Frontier Research Center funded by the U.S. Department of Energy (DOE), Office of Science, Basic Energy Sciences (BES), under Award \#DE-SC0023389 and by the US National Science Foundation (NSF; Grant Number MTM2-2025541, OIA-2242812). The authors acknowledge research support from Clemson University with a generous allotment of computation time on the Palmetto cluster.
\end{acks}

\bibliographystyle{ACM-Reference-Format}
\bibliography{sample-base}

\end{document}